\documentclass[letterpaper]{article} 
\usepackage{aaai23}  
\usepackage{times}  
\usepackage{helvet}  
\usepackage{courier}  
\usepackage[hyphens]{url}  
\usepackage{graphicx} 
\usepackage{natbib}  
\usepackage{caption} 
\usepackage{algorithm}
\usepackage{algorithmic}
\usepackage{amsmath}
\usepackage{newfloat}
\usepackage{listings}
\DeclareCaptionStyle{ruled}{labelfont=normalfont,labelsep=colon,strut=off} 
\floatstyle{ruled}
\newfloat{listing}{tb}{lst}{}
\floatname{listing}{Listing}
\title{Deep Generative Multimedia Children's Literature}
\author {
    Matthew L. Olson
}
\affiliations{
    Oregon State University\\
    olsomatt@oregonstate.edu
}

\begin{document}

\twocolumn[{%
\renewcommand\twocolumn[1][]{#1}%
\maketitle
\begin{center}
        \centering
        \vspace{-7mm}
        \includegraphics[width=0.85\linewidth]{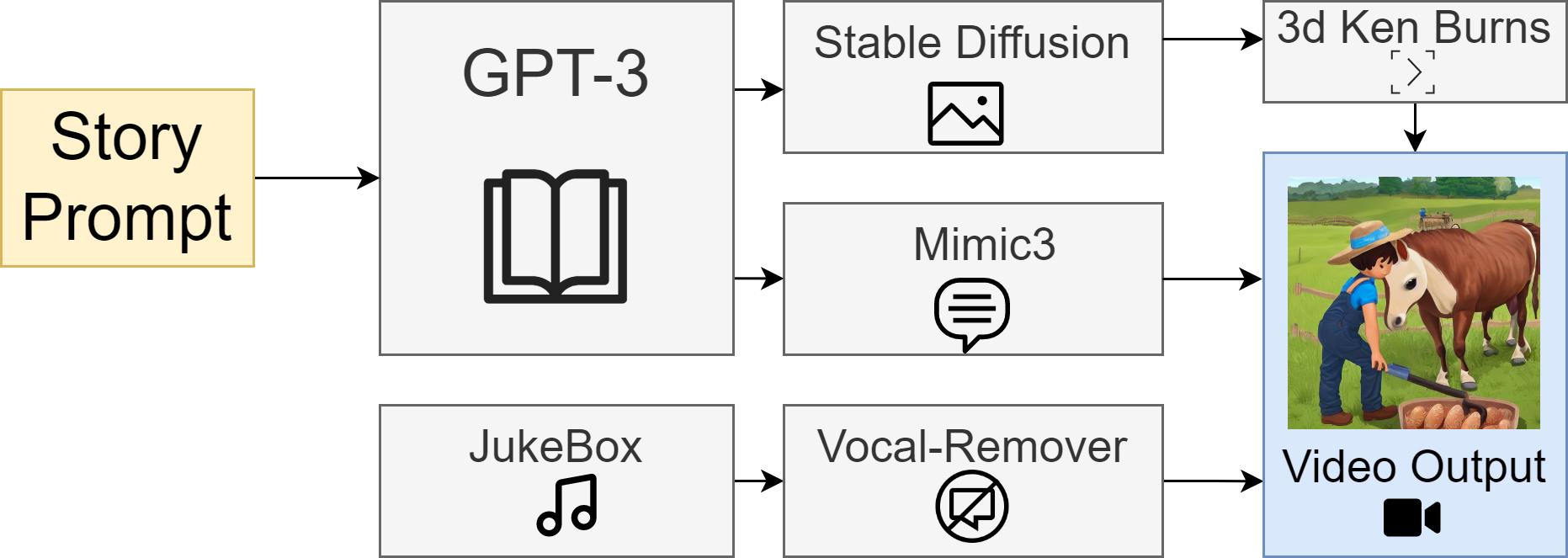}
        \captionof{figure}{
         A diagram of my  system architecture. An input prompt such as ``A children's story about a boy and a horse:'' results in a video of a read story with kids music. By leveraging a variety of publicly available Deep Learning models, I can create stories (GPT-3), images from text (Stable Diffusion), 3d effects (3d Ken Burns), voice from text (Mimic3), and kids music (JukeBox). The output video can be viewed at \url{https://youtu.be/YNvgHxaotQI}} 
        \vspace{2mm}
        \label{fig:diagram}
\end{center}%
}]

\begin{abstract}
Artistic work leveraging Machine Learning techniques is an increasingly popular endeavour for those with a creative lean. 
However, most work is done in a single domain: text, images, music, etc. In this work, I design a system for a machine learning created multimedia experience, specifically in the genre of children's literature. 
I detail the process for exclusively using publicly available pretrained deep neural network based models, I present multiple examples of the work my  system creates, and I explore the problems associated in this area of creative work.
\end{abstract}

\section{Introduction}
The popularity in Deep Learning (DL) based creative ventures continues to grow without any signs of slowing down. Unpredictable to many a decade ago, the achievements of DL models in a variety of creative domains are spectacular in their own right. 

In this work, I combine multiple publicly available DL models to create a fully automated system in the generation of multimedia entertainment. The framework I propose is general enough for any genre of entertainment, but I focus on the task of children's video literature production.

A single DL model can produce amazing results for one domain: creating artistic faces \citep{karras2020analyzing}, music \citep{dhariwal2020jukebox}, images from text \citep{ramesh2021zero}, and complete short stories \citep{brown2020language}. However, each of these models only functions in one or two domains. By combining a multitude of models, I create a system for generating children's video literature. However, this newfound creative achievement is not without major societal implications.

Internet videos targeted towards children gained much attention a few years ago when advertisers saw the bizarre and terrifying children videos on the internet \citep{maheshwari_2017}. While the policies of video content providing websites significantly changed, the underlying problem of video content being unsettling and novel still plagues user-content video applications. 
This work focuses on the creation of child-friendly content, but in order to shed light on the ease at which unscrupulous actors could create their own videos, I also explore how creating a fully automated system targeted for children will tend to result in disturbing content. 

The paper is structured as follows. Section $2$ provides a detailed explanation of the DL models used in my  system. Section $3$ explores the results generated by my  carefully constructed framework to produce child friendly content, as well as the unintended creation of unfriendly content. And I end with a discussion in section $4$ with an analysis of ethical implications of my  work, where I detail why I explore friendly and unfriendly content.
Examples of the full videos my  system created can be seen on my  dedicated YouTube channel for this project at \url{https://www.youtube.com/channel/UCnfcG7aaqHtldpzEWvqaIXQ} (no children were shown these videos in the making of this work). 

\section{Methods}
Here I describe the steps for combining publicly available pretrained DL models; a diagram depicting the process is shown in figure \ref{fig:diagram}.

\subsection{Language Model}
The central model of this system is GPT-3 \citep{brown2020language} 
\footnote{While model weights for GPT-3 are not available for download, an API is available for a nominal cost (less than $\$0.10$ per story). And a free version, OPT \cite{zhang2022opt}, does exist for those with compute power to run it.}.
GPT-3 is an auto-regressive large language model. Given a series of tokens $T = t_1,t_2,...,t_n$,  GPT-3 is trained to maximize the next token $P(t_{n+1}| T)$ which is the probability of token $t_{n+1}$ occurring after tokens $T = t_1,t_2,...,t_n$. By using an extremely large transformer-based deep neural network with 175 billion parameters, along with half a trillion tokens to train on, GPT-3 is able to convincingly generate text (of length $m$) $t_{n+1},...,t_{n+m}$ given a prompt $T$ by iteratively appending $t_{n+i}$ to $T$.

To generate an initial children's story $G$, I prompt GPT-3 with 
$T_G = $ ``The following is a children's story about a $X$ and a $Y$ :''
, where $X$ and $Y$ selected from a list of boy, girl, or some animal. GPT-3 is very effective at creating a simple and cohesive story that follow a logical flow. 

Let us define $T_{Di}$ as a re-prompt for GPT-3 with the entire story $G$, followed by the text: ``The story has pictures accompanying it. From the sentence $S_i$, here is what the picture looks like: '', where $S_i$ is one of the sentences from the story.  The output of GPT-3, given $T_{Di}$ is then a single description of an image associated with $S_i$, which can more formally be written as $D_i = GPT\text{-}3(G, T_{Di})$. I then save the output $D_i$ for of each of these sentence descriptions to be used for the text-to-image model.

\subsection{Text-to-Image Model}
Diffusion models are particularly interesting as they have been combined with language models to produce conditional image generation. By pretraining a language and vision encoder to jointly maximize the similarity of text and image embedding pairs, the language encoder learns a rich feature space which can be repurposed for text-conditioned image generation.
Stable Diffusion \citep{rombach2021highresolution} is the model of choice for text-to-image generation as it is trained heavily on artistic images and is freely available to download.
Stable Diffusion is composed of a text encoder and a denoising autoencoder such that for a given input image description $D$, it creates a text-conditioned image $I = StableDiffusion(D)$. As Stable Diffusion is trained on a large dataset, LAION \cite{desai2021redcaps}, of $5$ billion aesthetically pleasing text-image pairs from the internet, and it can generate a wide variety of artistic images given a prompt. 

\subsection{Static Video Enhancement}

\begin{figure*}[!htpb]
\centering
         
         \includegraphics[width=0.99\linewidth]{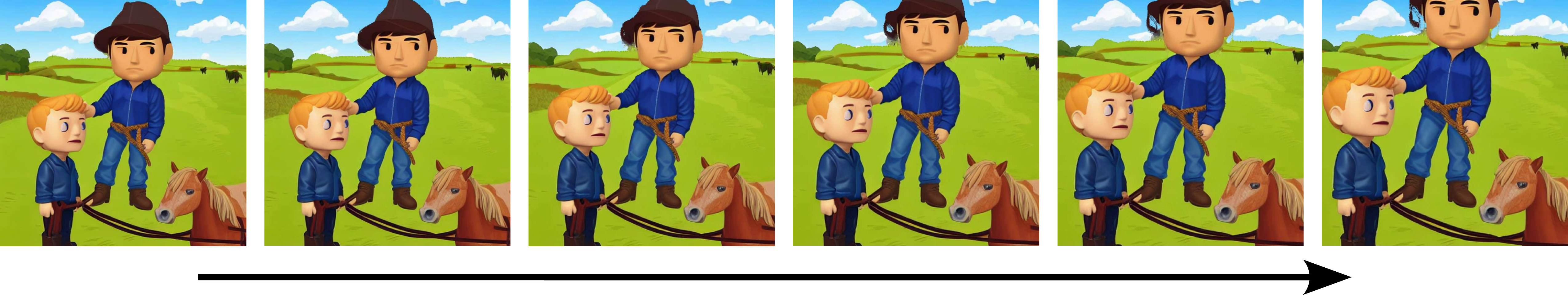}

        \caption{ An example of the 3d ken burns effect applied to one of the video scenes. The foreground is zoomed in, while the background contents (clouds in this example) are left unchanged.}

        \label{fig:3d_kenburns}
\end{figure*}

While static images for a video is a good replication of a physical storybook, I include another publicly available model to create a 3d Ken Burns Effects \citep{Niklaus_TOG_2019}. By combining a depth prediction DL model along with a context-aware synthesis network, this recent state-of-the-art model can take any static picture $I$, separate different elements from each other (like foreground objects and background objects), and create a zoom/pan/rotation video around those objects as if the image were 3d. By applying this model to the StableDiffusion images, I am able to create much more dynamic frames for the videos $K = 3d-KenBurns(I)$, adding an extra degree of freedom for each frame. An example of the effect applied to a scene is shown in figure \ref{fig:3d_kenburns}.

\subsection{Text-to-Speech Model}
To finish off the storytelling, I use the Mimic3 text-to-speech DL model \citep{gesling}. Mimic3 is based off VITS \cite{kim2021conditional}, which is a non-autoregressive model that directly predicts speech waveforms $W = Mimic3(G)$ from input text (i.e., a generated story $G$).  Mimic3 is an ideal piece of software as it has many choices for what voice to use, each of which is a good replica of a human voice. Mimic3 has many customization options, such as speech speed, which enables dynamic and easy application to the story. Other non-DL text-to-speech methods sound much more robotic and take away from the overall feel of the videos. 

\subsection{Music Models}
The emotional valence of video scenes are significantly impacted by accompanying music \cite{ansani2020soundtracks}. The choice of music for my proposed multimedia is crucial for setting the tone.  There are two major choices for selecting music: models that create raw sounds (wav files) or models that create notes (midi files). In this work, I investigate two DL models: JukeBox \citep{dhariwal2020jukebox} to create wav files and Perceiver-AR \citep{hawthorne2022general} to create midi files.

\noindent{\textbf{JukeBox}} leverages a VQ-VAE \cite{van2017neural} DL model to compress segments of a wav file into a significantly smaller latent representation, where it then auto-regressively predicts waveform segments (in the latent space) in order to create music with lyrics. While this model takes many hours to create one minute of music $\mathbf{M} = JukeBox()$, it has a wide variety of artists and genres to choose from including upbeat children's music artists. 

\textbf{Lyrical Removal Model}
One drawback of the JukeBox model is that it has no explicit option to disable lyrics. I use Vocal-Remover \cite{vocalremover} a which is based off of MMDenseNet \cite{takahashi2018mmdenselstm}, which takes an input song and returns two audio tracks: the singing and the instruments. And while this model does not perform perfectly with generated music, it does de-emphasize the singing in a given song. I therefore wrap JukeBox with this model to significantly soften the singing in my generated tracks. 

\noindent{\textbf{Perceiver-AR}} is the other option for creating music by modeling the physical process of playing an instrument as encoded by a midi file. Through training on publicly available classical music, any general auto-regressive DL model can easily be adapted for this type of data and the Perceiver-AR is a recent high quality DL model for such predictive modeling of music $\mathbf{M} = Perceiver\text{-}AR()$ by leveraging large contexts for prediction. The music produced by this model takes only seconds to create and sounds better than a randomly played piano-- and when it comes to ease of use goes for publicly available music generation models, the Perceiver-AR is significantly easier to setup to create music when compared to JukeBox. I include its use in my framework and explore the consequences of its use below.

An algorithmic description of the entire multimedia children’s literature generation is described in algorithm \ref{algo}.


\begin{algorithm}[!htpb]
    \caption{Psuedocode for Multimedia Generation}\label{algo}
    \begin{algorithmic}
        \STATE  $\mathrm{Generate\_Video} (X,Y)$
        \STATE $G = GPT\text{-}3(T_G(X,Y))$, Create story from prompt 
        \STATE $\mathbf{K} = []$, Initialize empty list of scenes 
        \FOR{Sentence $S_i$ in $G$}
            \STATE $D_i = GPT\text{-}3(T_{Di}(S_i))$, Image description 
            \STATE $I_i = StableDiffusion(D_i)$, Image 
            \STATE $K_i = 3dKenBurns(I_i)$, Dynamic scene 
            \STATE Append $K_i$ to $\mathbf{K}$
        \ENDFOR
        \STATE $W = Mimic3(G)$, Generate speech 
        \STATE $M = JukeBox()$, Generate music 
        \STATE $V = FFmpeg(\mathbf{K},W,M)$, Concatenate into video 
        \STATE Return $V$
    \end{algorithmic}
\end{algorithm}

\begin{figure}[!bt]
\centering
         
         \includegraphics[width=0.99\linewidth]{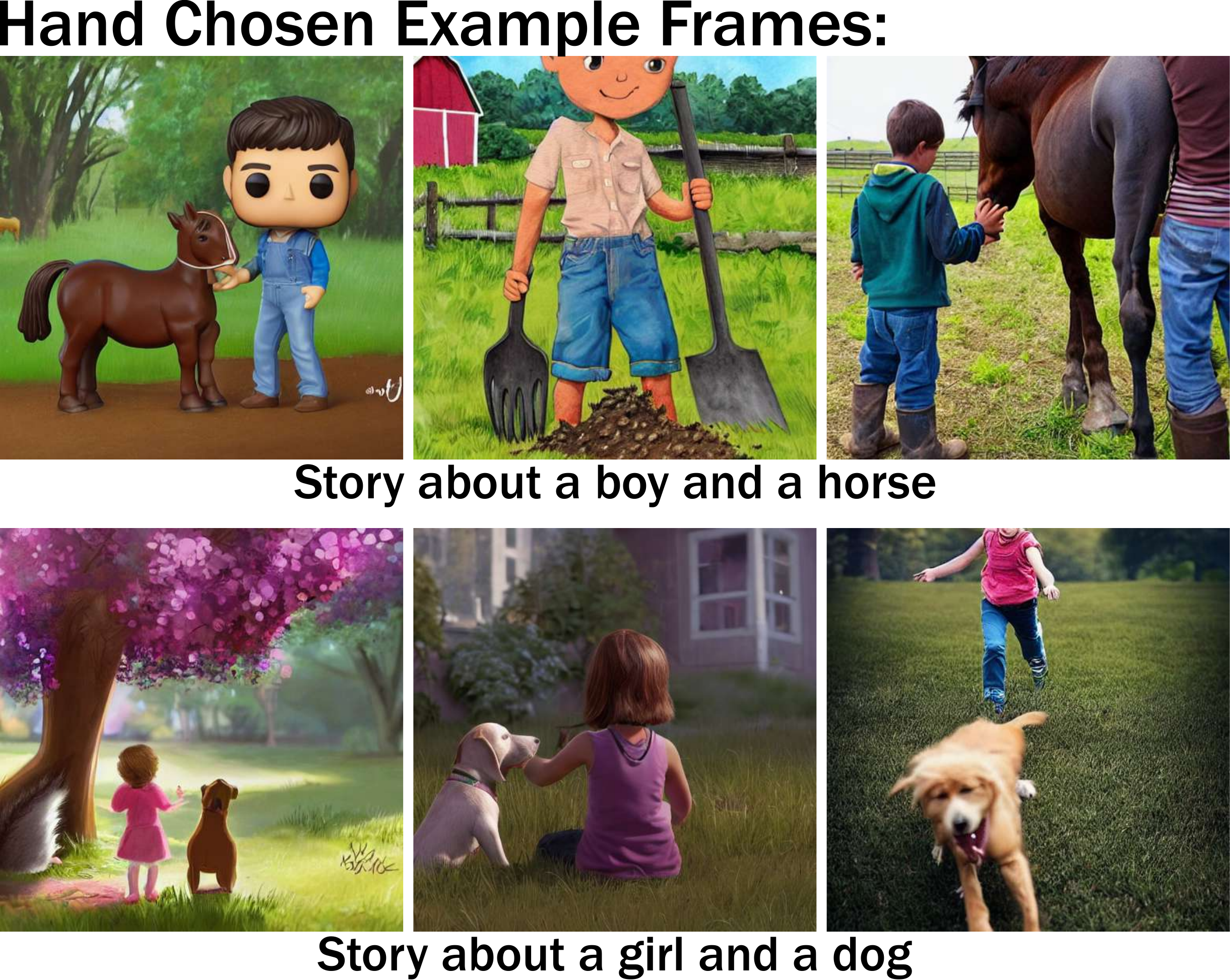}

        \caption{ Examples from two of our stories, where I hand-picked the frames for each scene. These images are generally cute and follow a consistent narrative flow for the given video.}

        \label{fig:cute}
\end{figure}
\begin{figure*}[!hpbt]
\centering
         
         \includegraphics[width=0.99\linewidth]{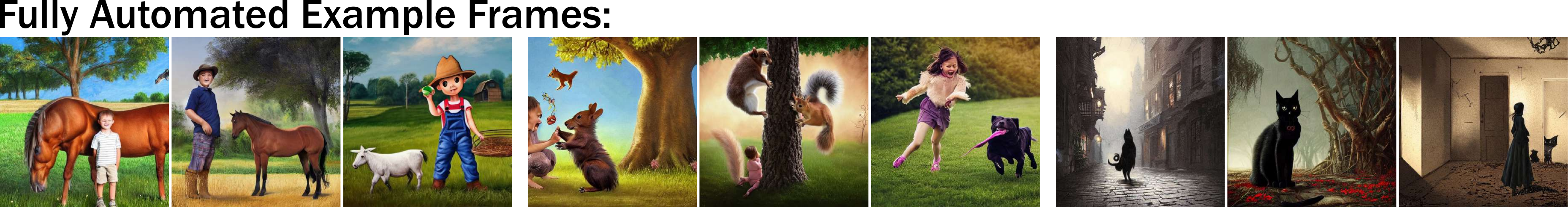}

        \caption{ Examples from using StableDiffusion in a fully automated way, where images are often unsettling and would need significant refinement in order to be ready for a children's media setting.}

        \label{fig:terrifying}
\end{figure*}

\subsection{Detailed Implementation}
\noindent{\textbf{GPT-3}} is very straightforward to use with their web-based interface. I use default parameters with the ``text-davinci-002'' model. This model is the most powerful and consistently provides excellent stories and image descriptions.

\noindent{\textbf{Mimic3}} has many options for speaker's voice, language, and speed. I use the free US English ``VCTK'' model, with a 2.1x speed scaling to slow the speech down.

\noindent{\textbf{JukeBox}} is trained on thousands of artists with many different genres. I select two popular children's music artists (The Wiggles and Raffi) and randomly sample a minute of unconditional music. The output wav file is then run through Vocal-Remover.

\noindent{\textbf{Perciever-AR}} needs to have its midi outputs converted into mp3 files, for which I use the window's operating system's default sound font file, along with VLC \cite{solutions2006vlc} to perform the conversion.

\noindent{\textbf{Stable Diffusion}} still requires some engineering of the input in order to generate interesting outputs that correspond to the desired image. For each sentence description of the story, I append a few words such as ``extremely detailed, textured, high detail, 4k'' in order to produce high resolution outputs, along with other major themes like ``Funko Pop''. Stable Diffusion can easily be run multiple times per input to create many different images to select from, in order to find images that are not unsettling, I generate $100$ different examples for every image description and pick one that I find cute and interesting. I also investigate a fully automated setup where only a single instance for each description is generated, finding it to have serious issues.

\noindent{\textbf{3d Ken Burns Effects}} provides an easy to use github implementation. I use the default zoom/unzoom, though other effects can easily be applied to my frames.

All output files from each of these above models are concatenated and combined using FFmpeg \cite{tomar2006converting}.


\section{Results}
Here I present two videos generated by my system, which can be seen on my YouTube channel. Example frames from both videos are shown in figure \ref{fig:cute}; while the style is not consistent scene to scene, these images seem appropriate for their context within the story.

The first video, \url{https://youtu.be/YNvgHxaotQI} is about a boy and a horse. The images are cohesive: showing a horse, boy, or farmer in the various scenes. The generated song by Raffi fits well, filling in pauses between sentences.
The second video, \url{https://youtu.be/PaivgKpS5Bo} is about a girl, a dog and a squirrel. Again, the scenes a overall consistent and the musical accompaniment of The Wiggles enhances the tone of the video. The story itself even contains a moral about boundaries. Overall, these videos are cute and something I could imagine a child watching on YouTube (though I do reiterate that no children were shown these videos at any point during the production of this work).

\subsection{Easy production, alarming results}
On the other hand, in my efforts to create end-to-end fully automated children's multimedia literature, I stumbled upon content that is the antithesis of my goals: moderately disturbing videos. Two aspects contribute to this problem: the music and uncurated images. The music produced Perceiver-AR takes only seconds to create and seems like a decent choice as it sounds better than a randomly played piano, however, it has the terrible result of sounding like it belongs in a horror film. Additionally, Stable Diffusion has significant quality issues when generating faces. Examples of this generation can be seen in figure \ref{fig:terrifying}.

I now present two additional videos from my YouTube channel, which have been described by viewers as "mildly terrifying". The first video, \url{https://youtu.be/AyeOWBPduM8} , uses the same story a boy and a horse. The video shows strange artefacts around horse generation and cannot accurately represent faces. The second video \url{https://youtu.be/sXEIMki9fZI} uses the same story about a girl, a dog and a squirrel. This video is a bit more frightening, with the images being more artistic and the animals much less defined than the horses of the previous video. 

\paragraph{Deep Generative Multimedia Horror Literature?}
Lastly, I present an alternative approach in the following video \url{https://youtu.be/JCe7KkVT3Ds}. Seeing how scary the music created by Perceiver-AR is, I prompted GPT-3 with "demon" and "cat", to get horror-based video, rather than a video for children. This story is significantly longer and not written in the third-person perspective. The first-person nature of the story means each frame has less characters to work with and makes more images that do not match the story. The long run-time of the story also shows the importance of the 3d ken burns effect, as using only the default zoom/unzoom becomes tedious by the end of this story. I hoped this direction of multimedia production would be more fruitful, but I leave multi-genre multimedia production for future work.

\section{Discussion}
\subsection{Ethical Implications}
I did not intend to create a children's video generator that produced disturbing content; the unsettling videos were an unintended consequence of blindly building a fully automated system. 
While I could have crafted a version of this work with no mention to these unfortunate videos, I believe it is my responsibility to disclose its accidental production.

I understand that some may find such a demonstration to be unacceptable, that belief (in my opinion) is similar to security through obscurity.
It is better to plainly show how even those with no ill intentions can accidentally generate this type of content.
My hope is that by including these examples, the creative machine learning community will be more aware of the unintentional side effects of publicly releasing their models. 
Furthermore, I hope to bring this content to light for parents, as the responsibility for children's video content consumption should fall on \textit{informed} parents who are aware of the dangers in user generated content. 

\subsection{Conclusion}
In this work, I proposed an system for creating children's multimedia literature. I showed that by combining a variety of pretrained Deep Learning models, full video stories are relatively easy to automatically generate. I discussed the practical application of each DL model and how each fit into the overall system. Lastly, I presented multiple different videos, analyzed their contents, and showed how my system can create good and bad children's videos. 

\bibliography{biblio.bib}

\end{document}